\documentclass[letterpaper]{article} 
\usepackage{aaai2026}  
\usepackage{times}  
\usepackage{helvet}  
\usepackage{courier}  
\usepackage[hyphens]{url}  
\usepackage{graphicx} 
\usepackage{natbib}  
\usepackage{caption} 
\usepackage{algorithm}
\usepackage{algorithmic}
\usepackage{amsmath}
\usepackage{newfloat}
\usepackage{listings}
\DeclareCaptionStyle{ruled}{labelfont=normalfont,labelsep=colon,strut=off} 
\floatstyle{ruled}
\newfloat{listing}{tb}{lst}{}
\floatname{listing}{Listing}
\usepackage{multirow}
\usepackage{booktabs}
\usepackage{color}
\usepackage{xcolor}
\usepackage{xspace}

\newcommand{\name}{SmartSight\xspace}
\usepackage{times}
\usepackage{helvet}
\usepackage{courier}
\usepackage{xcolor}

\title{SmartSight: Mitigating Hallucination in Video-LLMs Without Compromising Video Understanding via Temporal Attention Collapse}
\author {
    {Yiming Sun},
    {Mi Zhang},
    {Feifei Li},
    {Geng Hong}\footnotemark[2],
    {Min Yang}\thanks{Corresponding authors: Geng Hong and Min Yang.}
}
\affiliations {
    College of Computer Science and Artificial Intelligence, Fudan University, Shanghai, China\\
    ymsun24@m.fudan.edu.cn,  mi\_zhang@fudan.edu.cn, ffli23@m.fudan.edu.cn, \\ ghong@fudan.edu.cn, m\_yang@fudan.edu.cn
}

\usepackage{bibentry}

\begin{document}

\maketitle
\begin{abstract}
Despite Video Large Language Models (Video-LLMs) having rapidly advanced in recent years, perceptual hallucinations pose a substantial safety risk, which severely restricts their real-world applicability.
While several methods for hallucination mitigation have been proposed, they often compromise the model’s capacity for video understanding and reasoning.
In this work, we propose SmartSight, a pioneering step to address this issue in a training-free manner by leveraging the model’s own introspective capabilities.
Specifically, SmartSight generates multiple candidate responses to uncover low-hallucinated outputs that are often obscured by standard greedy decoding.
It assesses the hallucination of each response using the \textit{Temporal Attention Collapse} score, which measures whether the model over-focuses on trivial temporal regions of the input video when generating the response. 
To improve efficiency, SmartSight identifies the \textit{Visual Attention Vanishing} point, enabling more accurate hallucination estimation and early termination of hallucinated responses, leading to a substantial reduction in decoding cost.
Experiments show that SmartSight substantially lowers hallucinations for Qwen2.5-VL-7B by 10.59\% on VRIPT-HAL, while simultaneously enhancing video understanding and reasoning, boosting performance on VideoMMMU by up to 8.86\%.
These results highlight SmartSight’s effectiveness in improving the reliability of open-source Video-LLMs.

\end{abstract}

\section{Introduction}

Recently, Video Large Language Models (Video-LLMs) have demonstrated remarkable capabilities in visual perception, semantic understanding, and complex reasoning~\cite{hu2025videommmu,chattracker}. This has led to their widespread deployment in 
{real-world applications} 
such as autonomous driving and robotic manipulation~\cite{Zhao_Wang_Zhu_Chen_Huang_Bao_Wang_2025}.
Despite their impressive 
{performance}, 
Video-LLMs still face a major challenge: \textbf{perception hallucinations}. 
Specifically, they may misrepresent object attributes, fabricate nonexistent content, or overlook critical video segments.
This poses serious risks to the  {use} of Video-LLMs, especially in safety-critical scenarios.

\begin{figure}[t]
    \centering
    \includegraphics[width=0.95\linewidth]{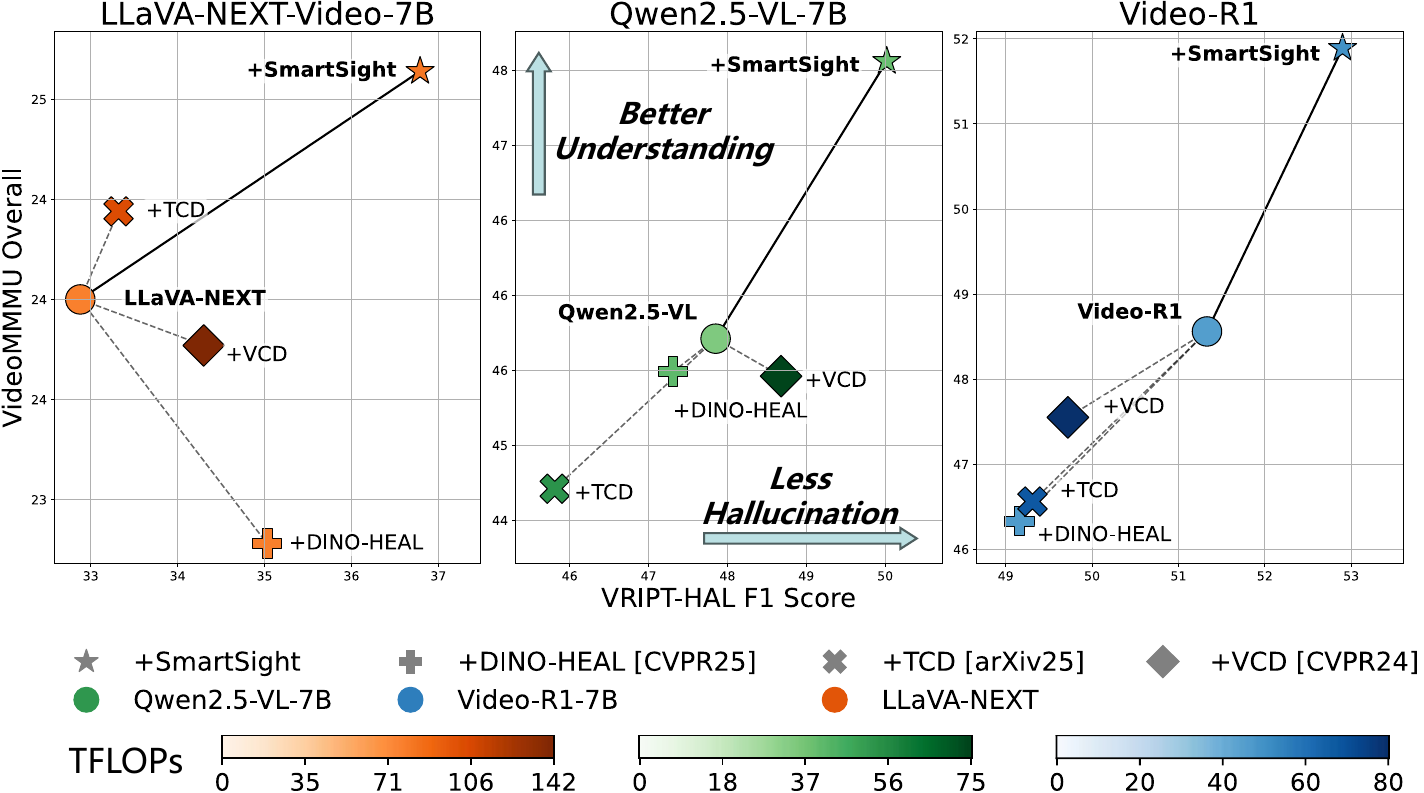}
    \caption{
Comparison between SmartSight and existing methods in terms of hallucinations suppression, video understanding and reasoning. Lighter colors indicate lower computational cost. Existing methods suffer from two main limitations: (1) inferior transferability across models, showing effectiveness only on LLaVA-NEXT-Video; and (2) reducing hallucinations at the cost of impaired  video understanding. SmartSight simultaneously suppresses hallucinations and enhances video comprehension, achieving a more favorable balance between accuracy and efficiency.
 }
    \label{fig:1}
\end{figure}

Although a few works have attempted to mitigate {perception} hallucinations in video perception by fine-tuning Video-LLMs~\cite{yang2024vript} or applying training-free methods such as Visual Contrastive Decoding~(VCD)~\cite{vcd}, we found that these methods significantly compromise the model’s essential ability for video understanding.
As shown in Figure~\ref{fig:1}, they suppress hallucinations only on LLaVA-Next-Video, while severely impairing the model’s ability to understand and reason.
One contributing factor is that inappropriate fine-tuning leads to catastrophic forgetting~\cite{li-etal-2024-revisiting}, while training-free methods are prone to misidentifying expected outputs as hallucinations and introduce significant computational overhead during inference~\cite{huo2025selfintrospective}.
In this work, we take an initial step toward bridging the gap between 
(1) mitigating perception hallucinations and (2) preserving understanding and reasoning capabilities in Video-LLMs.

\begin{figure}[t]
    \centering
    \includegraphics[width=1\linewidth]{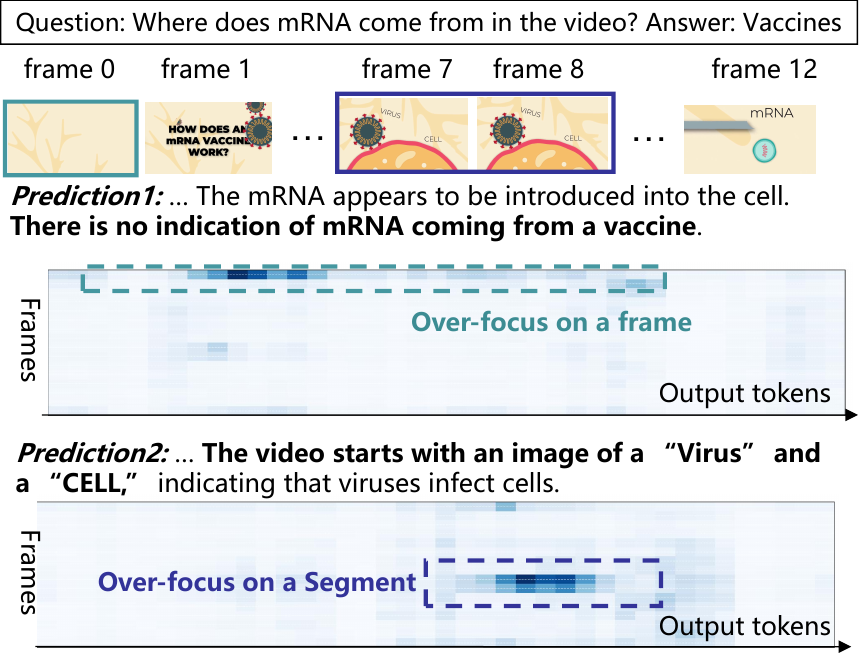}
    \caption{
Illustration of hallucination in Video-LLM outputs. Prediction 1 exhibits an over-reliance on the first frame, causing it to overlook crucial information in frame 12 and leading to an incorrect output. Prediction 2 concentrates excessively on visually similar segments (frames 7–8), which results in a misinterpretation of the video content.
    }
    \label{fig:2}
\end{figure}

\begin{figure}[t]
    \centering
    \includegraphics[width=1\linewidth]{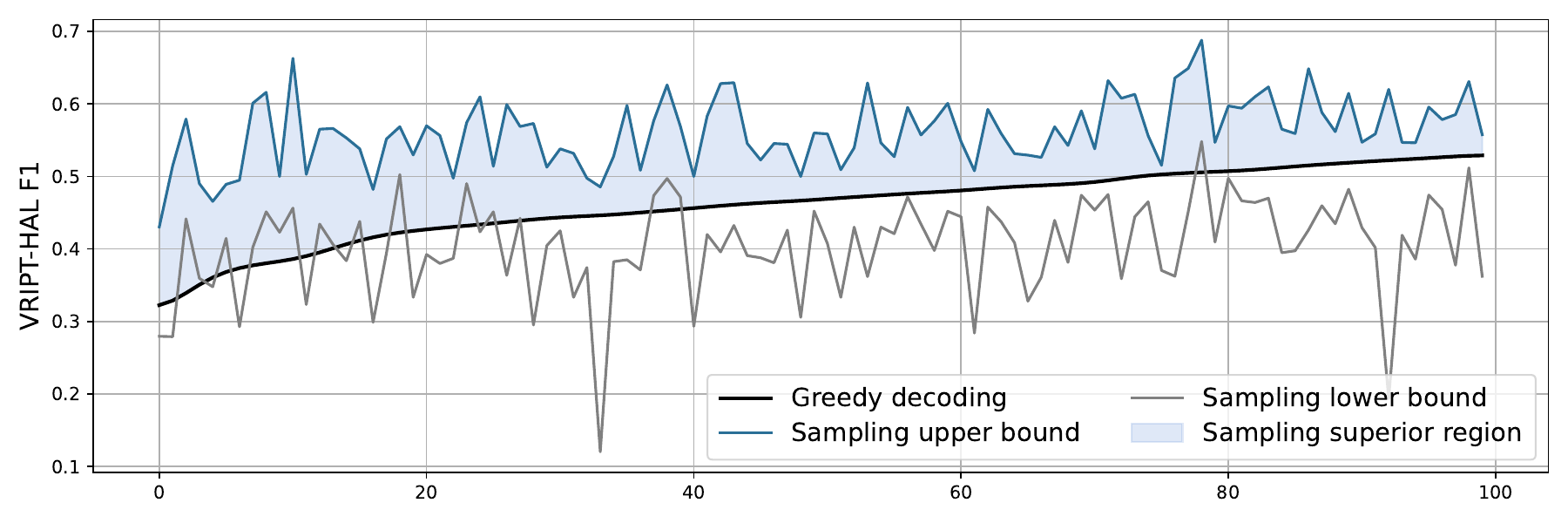}
    \caption{
 Comparison between greedy decoding and sampling  $N$=10 responses per query. We visualize 100 randomly selected responses from VRIPT-HAL. The black curve shows the hallucination level of responses generated by greedy decoding. The blue and gray curves indicate the least and most hallucinated among the sampled responses. The shaded region illustrates that sampling can produce responses with lower hallucination.
    }
    \label{fig:sampling_bound}
\end{figure}

Our method is grounded in several key 
observations.
We reveal that hallucinated outputs are associated with a collapse of attention onto a single frame or a semantically trivial segment -- {a phenomenon we term \textbf{Temporal Attention Collapse (TAC)}}. 
As shown in Figure~\ref{fig:2}, in \textit{Prediction 1}, model allocates excessive attention to a single frame, causing it to overlook other important parts of the video.
In \textit{Prediction 2}, it disproportionately focuses on a segment with limited variation in object appearance or motion, leading to a misinterpretation of event order. 
It is worth noting that over-focused frames or segments are not necessarily equivalent to key frames, as models need to leverage the video context to understand semantics before identifying key frames. One potential cause of TAC is the image bias inherent in Video-LLMs, where certain frames resemble images in the training data, thereby drawing disproportionate attention from the model~\cite{video-utr}.
Furthermore, sampling multiple responses uncovers the model’s capacity to generate outputs with \textit{fewer hallucinations}, which is often obscured by greedy decoding. As shown in Figure~\ref{fig:sampling_bound}, for each question, at least one of the ten sampled responses contains fewer hallucinations than the response generated with greedy decoding.
In addition, we observe a turning point during generation from which the model’s attention to visual information in the input video drops sharply -- referred to as \textbf{Visual Attention Vanishing (VAV) Point}. 
It marks a specific token after which attention to visual inputs has already substantially diminished.
The majority of attention to visual tokens is allocated before the VAV point. This reflects an inherent property of rotary positional encoding, where attention scores naturally decay as the relative distance between tokens increases~\cite{su2024roformer}.

Based on these findings, we propose a training-free method, dubbed \textit{\name}, that effectively mitigates hallucinations while enhancing 
the model's understanding and reasoning abilities.
\name generates multiple candidate responses to elicit less hallucinated outputs that are often obscured by standard greedy decoding~\cite{wan-etal-2025-reasoning}.
To facilitate the selection of a less hallucinated response, we propose the \textit{Temporal Attention Collapse} Score (TAC~score) as an introspective metric for evaluating the hallucination severity of each candidate.
Building on the observation that hallucinated outputs tend to over-focus on semantically trivial video segments or frames, the TAC~score evaluates hallucinations without relying on external models.
To improve efficiency, we further assess response quality at the proposed \textit{Visual Attention Vanishing} (VAV) Point, which enables 
accurate estimation of hallucination severity via the TAC~score
without waiting for full generation.
Responses with low TAC~score at this point are terminated early, reducing the computational overhead in the decoding stage by up to 79.6\%, with negligible impact on overall response quality.
Experimental results demonstrate that our method consistently enhances the understanding and reasoning capabilities of 10 diverse Video-LLMs. With  \name , Qwen2.5-VL-7B achieves performance competitive with Qwen2.5-VL-32B and proprietary Gemini 1.5 Pro. As the model size increases, \name exhibits a favorable scaling property not observed in existing methods.
Overall, these findings highlight \name as a robust and plug-and-play solution for improving the reliability of Video-LLMs in more challenging scenarios, such as harmful video detection, video-based jailbreak attack defense, and video quality assessment.

In summary, our main contributions are threefold:
\begin{itemize}
     \item We reveal that current hallucinations mitigation methods may unintentionally impair the semantic understanding and reasoning capabilities of Video-LLMs.
    \item We uncover the Temporal Attention Collapse phenomenon in hallucinated responses, characterized by excessive attention to trivial temporal regions and consistently observable across a variety of Video-LLMs.
    \item We propose \name, a pioneering step to alleviate perception hallucinations and improve video understanding and reasoning capabilities in a training-free manner, with impressive scalability and compatibility across a wide range of Video-LLMs. 
\end{itemize}

\section{Related Works}
\begin{figure*}[t]
    \centering
    \includegraphics[width=0.99\linewidth]{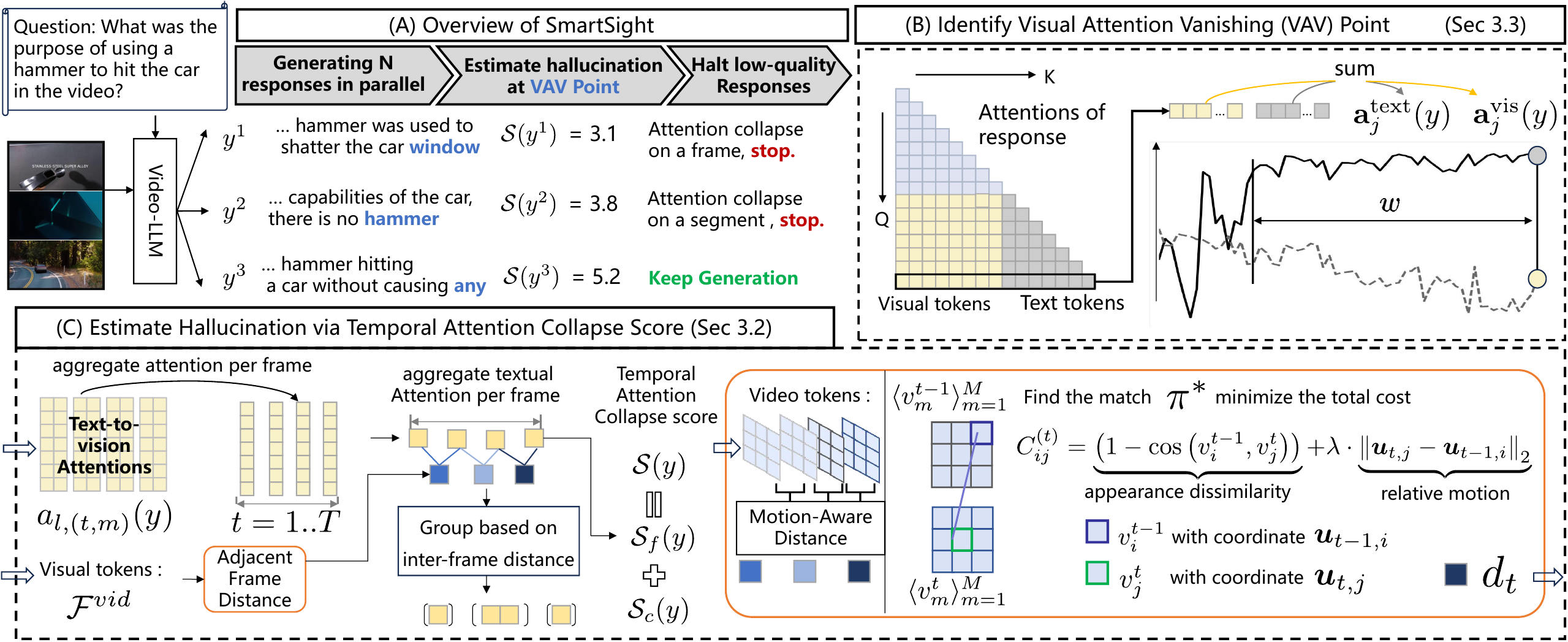}
    \caption{  Overview of the proposed SmartSight.  
Given an input video and a textual query, SmartSight generates $N$ responses in parallel. During generation, it dynamically detects the Visual Attention Vanishing Point and estimates hallucination severity using the proposed Temporal Attention Collapse Score. Only one high-quality candidate is retained for continued generation, thereby achieving a favorable balance between efficiency and effectiveness.
    }
    \label{fig:main}  
\end{figure*}
\subsection{Video Large Language models}
In recent years, Video Large Language Models~(Video-LLMs) have made rapid progress, leading to a wide range of real-world applications~\cite{you2025timesoccer,10094642,chen2025motionzerozeroshotmovingobject}. 
Video-LLMs encode video frames using a vision encoder, concatenate their embeddings across frames, and project them into the LLM’s linguistic embedding space via a cross-modal adapter~\cite{qwen2.5vl}.
This design enables Video-LLMs to perceive video  and understand its semantics~\cite{hu2025videommmu}. 
To further improve their capacity for complex reasoning, recent efforts such as Video-R1~\cite{video-r1}, VideoChat-R1~\cite{li2025videochat}, and TinyLLaVA-Video-R1~\cite{zhang2025tinyllava} have introduced reinforcement learning and test-time scaling strategies, inspired by the success of large-scale reasoning models like DeepSeek-R1~\cite{guo2025deepseek}.
Our method further improves the understanding and reasoning capabilities of these models by suppressing hallucinations.

\subsection{Evaluation for Hallucination in Video-LLMs}
Despite their impressive performance, Video-LLMs are more prone to hallucinations than image-based models, as the temporal dimension introduces unique challenges.
In addition to hallucinations in image tasks, Video-LLMs may misidentify actions, confuse the sequence of events, omit essential segments, or  fabricate scenes that are not present in the  video.
Existing hallucination evaluation methods for Video-LLMs primarily fall into two categories: VQA-based~\cite{zhang2025eventhallusion} and caption-based approaches~\cite{yang2024vript}. VQA-based methods formulate multiple-choice questions that query the model about the presence of plausible yet incorrect objects or events in a given video, thereby assessing whether the model hallucinates content not grounded in the visual input.

While straightforward, Video-LLMs tend to exhibit more hallucinations in open-ended text generation tasks (e.g., video captioning) than in multiple-choice verification~\cite{rawal2025argus}.  To address this limitation, caption-based evaluation methods quantify hallucination occurrence by comparing model-generated captions with human-provided references~\cite{yang2024vript}.
However, these methods typically fail to account for the impact of hallucination mitigation strategies on the model’s utility, such as its capacity for semantic understanding and reasoning. Our proposed method demonstrates competitive results under both evaluation protocols, effectively reducing hallucinations and enhancing the model’s usability.

\subsection{Mitigation for Hallucination in Video-LLMs}
Existing approaches for mitigating hallucinations in Video-LLMs can be broadly categorized into training-based and training-free methods.
Training-based methods, such as Vriptor~\cite{yang2024vript}, retrain models with densely annotated video captions to enhance fine-grained visual perception.
MASH-VLM~\cite{Bae_2025_CVPR} and Vista-LLaMA~\cite{ma2023vista} modify the self-attention mechanism to better capture temporal dependencies, which requires model retraining.
PaMi-VDPO~\cite{ding2025pami} constructs preference data via data augmentation and applies Direct Preference Optimization to fine-tune Video-LLMs.
While effective, these methods involve architectural changes and fine-tuning, which limit the generalizability of the fine-tuned model across domains such as mathematical reasoning.

To overcome these limitations, recent efforts have explored training-free hallucination mitigation.
Temporal Contrastive Decoding~\cite{zhang2025eventhallusion}, inspired by Visual Contrastive Decoding~\cite{vcd}, perturbs the input video to amplify potential hallucinations, then compares the outputs before and after perturbation to identify hallucinated tokens. However, improper perturbations may misidentify correct content as hallucinations.
DINO-HEAL~\cite{Li_2025_CVPR} addresses this by leveraging DINOv2~\cite{oquab2023dinov2} saliency maps to highlight key regions and suppress attention to background content.
However, we find that both training-based and training-free methods may undermine model usability. Furthermore, their effectiveness does not generalize well to reasoning-oriented models such as Video-R1~\cite{video-r1}. 
\section{Methodology}

\noindent \textbf{Overview:} 
As shown in Figure~\ref{fig:main} (A), \name elicits the model’s self-introspective capabilities to generate outputs with fewer hallucinations by sampling multiple responses. To evaluate the faithfulness of each response to the video content, we introduce Temporal Attention Collapse (TAC) Score (Sec.~\ref{sec:TAC}, Figure~\ref{fig:main} (C)), which assesses hallucination severity without relying on external models. 
We propose the Visual Attention Vanishing (VAV) Point in Sec.~\ref{sec:VAV} (Figure~\ref{fig:main} (B)), which facilitates early estimation of hallucination severity before full generation. Responses with higher hallucination levels are terminated in advance, thereby improving overall efficiency.
For clarity, we first introduce the notations used throughout this section in Sec.~\ref{sec:preliminary}.

\subsection{Preliminary}
\label{sec:preliminary}
\noindent \textbf{Input Formulation.}
Video-LLMs\footnote{Our study focuses on visual information, excluding subtitles or audio tracks.} receive a sequence of $T$ frames $\mathcal{V} = \langle f^t \rangle_{t=1}^{T}$ together with a textual question $\mathcal{X} = \langle x_l \rangle_{l=1}^{L}$ as input. 
They firstly divide each frame $f^t$ into $M$ patches $f^t = \langle p_m^t \rangle_{m=1}^{M}$, where each patch is then encoded into a visual token ${v}_m^{t}$. 
Consequently, a video with $T$ frames yields $T \times M$ visual tokens in total. 
The textual input is also encoded into textual tokens: $\mathcal{F}^{text} = \langle t_l \rangle_{l=1}^{L}$. 
The visual tokens $\mathcal{F}^{vid} = \langle  {v}_{m}^{t} \rangle_{m=1,t=1}^{M,\,T}$ and the text tokens are concatenated to form the full input $\mathcal{F}^ = \mathcal{F}^{vid} \cup \mathcal{F}^{text}$ to the language model.

\noindent \textbf{Stochastic Sampling.}
Let $p$ be a language model. Given input $\mathcal{F}$, the model generates a response $y = (y_1, y_2, \ldots, y_l)$ auto-regressively, where $l$ is the number of tokens. At each decoding step $k$, the next token is sampled from:
$y_{k+1} \sim p(\cdot \mid \mathcal{F}, y_{\leq k})$.
Unlike standard greedy decoding, which deterministically selects the token with the highest probability, sampling from the distribution promotes diversity and avoids linguistic bias.
We denote the full response sampling process as $y \sim p(\cdot \mid \mathcal{F})$. We then sample $N$ i.i.d. responses: $y^{1},\dots,y^{N} \sim p(\cdot \mid \mathcal{F} )$, where $y^{n}_i$ is the $i$-th token of the $n$-th response~\cite{renze-2024-effect}. 

\subsection{Self-Introspective Hallucination Estimation via Temporal Attention Collapse}
\label{sec:TAC}
In this section, we leverage the observed phenomenon of Temporal Attention Collapse to evaluate hallucinations solely on the model’s internal attention distribution.
Given a generated response $y = (y_1, \dots, y_L)$ with $L$ tokens, we compute the average attention received by each frame during the generation of $y$, denoted as: $
\mathbf{a}_t(y)=\frac{1}{ML} \sum_{m=1}^{M} \sum_{j=1}^{L}  a_{j,(t,m)}{(y)}$.
To simplify notation, we refer to $\mathbf{a}_{j,(t,i)}(y)$ and $\mathbf{a}_t(y)$ as $\mathbf{a}_{j,(t,i)}$ and $\mathbf{a}_t$, respectively, throughout this section.
 Next, we evaluate hallucination severity from two aspects: Frame-level Collapse and Segment-level Collapse.

 \noindent \textbf{Frame-level Collapse. } To quantify whether attention is disproportionately concentrated on a single frame, we define the Frame-level Collapse Score~$\mathcal{S}_f(y)$ for the response  $y$  :
\begin{align} 
\mathcal{S}_f(y) = -\sum_{t=1}^{T} \hat{\mathbf{a}}_t \log \hat{\mathbf{a}}_t, \quad
\hat{\mathbf{a}}_t = \frac{\mathbf{a}_t}{\sum_{t'=1}^T \mathbf{a}_{t'}}
\end{align}
The entropy of the normalized attention scores is used to assess the distribution of attention across frames. A lower $\mathcal{S}_f(y)$ indicates a sharp concentration of attention on one or a few isolated frames, suggesting a stronger tendency toward Frame-level Collapse.

 \noindent \textbf{Segment-level Collapse. } To assess whether attention is disproportionately allocated to  segments with little variation in object appearance or motion, we introduce the Segment-level Collapse Score $\mathcal{S}_c(y)$. 
We begin by analyzing inter-frame similarity based on video tokens $\mathcal{F}^{\text{vid}}$  to identify the trivial video segments. However, a direct use of cosine similarity between adjacent frames struggles to account for motion and  fine-grained appearance differences.
To address the issue, we define a Motion-Aware Cost Matrix:
\begin{align}
    C_{ij}^{(t)} =  \left(1 - \cos\left(v^{t-1}_{i}, v^{t}_{j} \right) \right) +  \left\| \boldsymbol{u}_{t,j} - \boldsymbol{u}_{t-1,i} \right\|_2 
\end{align}
 Here, ${v}_{j}^{t}$ denotes the visual embedding of the $j$-th patch in frame $t$, and $\boldsymbol{u}_{t,j}$ is its normalized  spatial coordinate.
The first term measures appearance dissimilarity between patch $i$ in frame $t{-}1$ and patch $j$ in frame $t$ via cosine distance, while the second term captures their relative motion based on spatial displacement.
Using this cost matrix,  we define the inter-frame distance $d_t$ as the minimum total matching cost under a one-to-one correspondence between patches:
\begin{align}
d_t = \sum_{i=1}^{N} C^{(t)}_{i, \pi^{*}(i)}, \quad \pi^*= \arg\min_{\pi \in \mathcal{B}_N} \sum_{i=1}^{N} C^{(t)}_{i, \pi(i)}
\end{align}
where $\mathcal{B}_N$  denotes   all possible one-to-one matchings between patches in frame $t{-}1$ and frame $t$. 

A higher $d_t$ suggests a larger degree of dissimilarity between the two frames.
We segment the video into $K$ temporally contiguous parts $\{s_k|k\in\{1, \dots, K\}\}$ by selecting frame $t$ as a boundary if its inter-frame distance $d_t$ exceeds a predefined threshold, i.e., $d_t > \gamma$. 
For each segment, the attention it receives is aggregated and normalized as $\hat{\mathbf{a}}_{s_k}{(y)} = {\mathbf{a}_{s_k}(y)}/{\sum_{\ell=1}^{K} \mathbf{a}_{s_\ell}{(y)}}$, where $\mathbf{a}_{s_k}{(y)} = \sum_{t \in {S}_k} \mathbf{a}_t{(y)}$.
The Segment-level Collapse Score $\mathcal{S}_c(y)$ is defined as the entropy of the resulting attention distribution:
$
\mathcal{S}_c(y) = - \sum_{k=1}^{K} \hat{\mathbf{a}}_{s_k}{(y)} \log \hat{\mathbf{a}}_{s_k}{(y)}
$. Finally, the overall Temporal Attention Collapse Score is defined as: 
\begin{align}
    \mathcal{S}(y) = \mathcal{S}_f(y) + \mathcal{S}_c(y)  \label{eq:tac}
\end{align}
A lower $\mathcal{S}(y)$ indicates that the model’s attention is more sharply concentrated on isolated frames and semantically trivial segments, suggesting a higher degree of hallucination.
\begin{table*}[t!]
\setlength{\tabcolsep}{4pt} 
\centering

\begin{tabular}{l@{\hskip 4pt}lc@{\hskip 4pt}c|l|lc@{\hskip 4pt}c@{\hskip 4pt}c|lc@{\hskip 4pt}c@{\hskip 4pt}c}
\toprule
                     & \multicolumn{3}{c}{VRIPT-HAL} & \multicolumn{1}{c}{EventHallu.}   & \multicolumn{4}{c}{Video-MME}    & \multicolumn{4}{c}{Video-MMMU}        \\
                     & F1~($\Delta$)     & R      & P      & \multicolumn{1}{c|}{Acc~($\Delta$)   }         & Overall~($\Delta$)    & {\small Long } & {\small Med.} & {\small Short}& Overall~($\Delta$)    & {\small Perc.} & {\small Compr.} & {\small Adap.} \\
 \midrule
Qwen2.5-VL &47.9 &{\small 35.6  }& {\small 75.0  }& 60.4 &54.8 &{\small 45.4  }& {\small 52.5  }& {\small 66.5  }& 45.7 &{\small 56.0  }& {\small 50.0  }& {\small 31.0 } \\
{\small +Dinoheal }  &47.3 (-0.54)&{\small 35.0  }& {\small 75.9  }& 56.8 (-3.53)&54.6 (-0.19)&{\small 47.4  }& {\small 51.9  }& {\small 64.5  }& 45.5 (-0.22)&{\small 58.3  }& {\small 48.0  }& {\small 30.1 } \\
{\small +TCD }  &45.8 (-2.04)&{\small 33.4  }& {\small 76.2  }& 61.4 (+1.01)&54.7 (-0.09)&{\small 44.1  }& {\small 53.1  }& {\small 66.8  }& 44.7 (-1.00)&{\small 54.7  }& {\small 47.7  }& {\small 31.6 } \\
{\small +VCD }  &48.7 (+0.83)&{\small 36.0  }& {\small 75.4  }& 61.9 (+1.51)&54.8 (-0.02)&{\small 45.1  }& {\small 52.3  }& {\small 67.0  }& 45.5 (-0.25)&{\small 59.3  }& {\small 46.5  }& {\small 30.5 } \\
{\small +SmartSight }  &50.0 (\textbf{+2.17})&{\small 38.5  }& {\small 73.8  }& 63.1 (\textbf{+2.78})&56.2 (\textbf{+1.39})&{\small 49.3  }& {\small 53.2  }& {\small 66.1  }& 47.6 (\textbf{+1.85})&{\small 59.7  }& {\small 51.3  }& {\small 31.7 } \\
 \midrule
Video-R1 &51.3 &{\small 40.3  }& {\small 71.8  }& 62.6 &59.8 &{\small 50.8  }& {\small 59.7  }& {\small 69.0  }& 48.6 &{\small 64.0  }& {\small 47.3  }& {\small 34.3 } \\
{\small +Dinoheal }  &49.2 (-2.17)&{\small 37.8  }& {\small 71.3  }& 62.6 (+0.01)&59.2 (-0.59)&{\small 50.1  }& {\small 58.6  }& {\small 69.0  }& 46.3 (-2.23)&{\small 62.0  }& {\small 46.3  }& {\small 30.7 } \\
{\small +TCD }  &49.3 (-2.02)&{\small 37.6  }& {\small 72.9  }& 64.9 (+2.27)&59.7 (-0.07)&{\small 50.1  }& {\small 59.1  }& {\small 70.0  }& 46.6 (-2.00)&{\small 63.7  }& {\small 44.3  }& {\small 31.7 } \\
{\small +VCD }  &49.7 (-1.61)&{\small 38.1  }& {\small 72.9  }& 65.4 (+2.78)&59.8 (+0.03)&{\small 50.2  }& {\small 59.9  }& {\small 69.4  }& 47.5 (-1.01)&{\small 64.0  }& {\small 45.7  }& {\small 33.0 } \\
{\small +SmartSight }  &52.9 (\textbf{+1.57})&{\small 43.0  }& {\small 69.7  }& 65.5 (\textbf{+2.89})&60.2 (\textbf{+0.41})&{\small 51.1  }& {\small 59.1  }& {\small 70.4  }& 51.9 (\textbf{+3.33})&{\small 66.0  }& {\small 50.7  }& {\small 39.0 } \\
 \midrule
LLaVA-NEXT &32.9 &{\small 22.0  }& {\small 69.0  }& 53.4 &28.8 &{\small 25.9  }& {\small 32.1  }& {\small 28.4  }& 24.0 &{\small 26.7  }& {\small 18.9  }& {\small 26.3 } \\
{\small +Dinoheal }  &35.0 (+2.15)&{\small 24.3  }& {\small 66.6  }& 53.9 (+0.51)&28.3 (-0.50)&{\small 28.1  }& {\small 28.1  }& {\small 28.6  }& 22.8 (-1.19)&{\small 27.0  }& {\small 18.0  }& {\small 23.3 } \\
{\small +TCD }  &34.3 (+1.42)&{\small 23.2  }& {\small 69.2  }& 55.7 (+2.36)&31.1 (+2.39)&{\small 27.9  }& {\small 30.9  }& {\small 34.7  }& 23.8 (-0.20)&{\small 27.3  }& {\small 16.7  }& {\small 27.3 } \\
{\small +VCD }  &33.3 (+0.44)&{\small 22.5  }& {\small 68.7  }& 53.1 (-0.25)&30.6 (+1.80)&{\small 27.6  }& {\small 30.6  }& {\small 33.6  }& 24.4 (+0.47)&{\small 23.7  }& {\small 20.3  }& {\small 29.3 } \\
{\small +SmartSight }  &36.8 (\textbf{+3.91})&{\small 26.0  }& {\small 67.6  }& 55.8 (\textbf{+2.43})&34.0 (\textbf{+5.22})&{\small 31.2  }& {\small 34.8  }& {\small 35.9  }& 25.1 (\textbf{+1.17})&{\small 27.1  }& {\small 22.0  }& {\small 26.3 } \\
   \bottomrule
\end{tabular}
\caption{
Comparison of state-of-the-art mitigation methods including DINO-HEAL (CVPR'25), TCD (arXiv’25), and VCD (CVPR’24) on three Video-LLMs: Qwen2.5-VL-7B, Video-R1, and LLaVA-NEXT-Video-7B. VRIPT-HAL and EventHallusion~(binary QA subset) evaluate hallucination suppression, while VideoMME and VideoMMMU assess understanding and reasoning capabilities. Higher scores indicate better performance. The most significant improvements are highlighted in \textbf{bold}.
}

\end{table*}
\subsection{ Efficient Generation via Early Stopping at Visual Attention Vanishing Point }
\label{sec:VAV}

We begin by generating $N$ responses in parallel, denoted as $Y = \{ y^{1}, y^{2}, \ldots, y^{N} \}$, where each $y^{n} \sim p(\cdot \mid \mathcal{F})$ is
a response independently sampled from the model.
This helps the model to explore diverse reasoning paths and break free from over-reliance on its biased prior, resulting in outputs that better align with the visual content.

To improve efficiency, we assess the quality of each sampled response based on its partial output truncated at the Visual Attention Vanishing (VAV) Point, instead of completing the entire generation. The VAV Point marks the stage when attention to visual tokens has substantially diminished.
To identify this point on-the-fly during the generation of n-th  response $y^{n}$, we compute $ \mathbf{a}_j^{\text{vis}}(y^{n}) $ and $\mathbf{a}_j^{\text{text}}(y^{n})$, which denote the attention that the $j$-th token assigns to visual tokens and preceding text tokens, respectively.
\begin{align}
  \mathbf{a}_j^{\text{vis}}(y^{n}) &= \sum_{t=1}^{T} \sum_{m=1}^{M} a_{j,(t,m)}({y^{n}}),  \notag \\ 
  \mathbf{a}_j^{\text{text}}(y^{n}) &= \sum_{l=1}^{j-1} a_{j,l}({y^{n}})  
\end{align}
Here, $a_{j,(t,m)}(.)$  denotes the attention weight from the $j$-th output token to the $m$-th visual token $v^t_m$ in $t$-th frame, averaged across all attention heads and layers. Similarly,  $a_{j,l}(.)$ represents the averaged attention from the $j$-th token  to the $l$-th generated token.

We define the VAV Point of response $y^{n}$ as the earliest generation step $j_{\text{vav}}(y^n) $ at which the model’s attention to visual tokens remains low for $w$ consecutive steps:
{\small
\begin{align}
j_{\text{vav}}(y^n) = \min \left\{ j \,\middle|\, \frac{\mathbf{a}_{k}^{\text{vis}}(y^{n})}{\mathbf{a}_{k}^{\text{text}}(y^{n})} < \alpha,\; \forall k \in [j - w + 1, j] \right\}
\end{align}
}
The ratio ${\mathbf{a}_{k}^{\text{vis}}(y^{n})}/{\mathbf{a}_{k}^{\text{text}}(y^{n})}$ indicates the model’s relative attention to visual input versus previously generated text at step $k$.
By requiring the ratio to remain below the threshold $\alpha$ for $w$ consecutive tokens, we ensure the VAV Point is not triggered by short-term fluctuations in attention.
We then compute the TAC Score (Eq.~\ref{eq:tac}) for each sampled response truncated at its VAV Point and select the top-$1$ candidates:
\begin{align}
 n^* &=  \arg\max_{n} \left( \mathcal{S}(y^{n}_{1:j_{\text{vav}}{(y^n)}}) \,\middle|\, n = 1,\ldots,N \right) \end{align}
The selected candidate response $y^{ n^*}$ is then allowed to complete generation and is used as the final output.By terminating the generation of suboptimal candidates early,we achieve a favorable trade-off between performance and efficiency.

\section{Experiments}

\subsection{Experimental Settings}
\noindent \textbf{Implementation Details.}
We use temperature sampling~\cite{renze-2024-effect} with a temperature of 0.7 to generate $N{=}10$ candidate responses before early stopping. Hyperparameters $\alpha$ and $w$ are set to 1.2 and 10, respectively. All methods use 32 sampled frames with a fixed resolution of 392 for fair comparison. 

\noindent \textbf{Benchmarks.}
We assess hallucination mitigation on the caption-based VRIPT-HAL~\cite{yang2024vript} and VQA-based EventHallusion~\cite{zhang2025eventhallusion} benchmarks.
To evaluate video understanding and reasoning capabilities, we utilize two large-scale, long-form, and open-domain benchmarks: Video-MME~\cite{Fu_2025_CVPR} and Video-MMMU~\cite{hu2025videommmu}.
For the VQA task, all methods adopt the same chain-of-thought (CoT) prompt and the official evaluation code.
Following standard practice for isolating the contribution of the visual modality~\cite{Fu_2025_CVPR}, we do not include subtitles during evaluation.

\noindent \textbf{Baseline Methods.}
We compare \name against three representative  hallucination mitigation methods: two approaches specifically designed for video understanding, TCD~\cite{zhang2025eventhallusion} and DINO-HEAL~\cite{Li_2025_CVPR}, as well as VCD~\cite{vcd}, a method initially developed for images but transferable to video tasks. All methods are implemented using the default settings as described in their original papers.

\begin{table}[t!]

\setlength{\tabcolsep}{4pt}

\begin{tabular}{cccccc}
\toprule
\multicolumn{2}{c}{Qwen2.5-VL} & \multicolumn{1}{c}{Base} & \multicolumn{1}{c}{TCD} & \multicolumn{1}{c}{VCD} & \multicolumn{1}{c}{SmartSight} \\
\midrule
\multirow{2}{*}{3B} &  {\small VRIPT-HAL} & 28.3 & 28.8 & 29.0 & 30.7 \\
 & {\small Video-MMMU} & 37.3 & 34.9 & 35.0 & 38.3 \\
 \midrule
\multirow{2}{*}{7B} & {\small VRIPT-HAL} & 47.9 & 45.8 & 48.7 & 50.0 \\
 & {\small Video-MMMU} & 45.7 & 44.7 & 45.5 & 47.6 \\
\midrule
\multirow{2}{*}{32B} &{\small VRIPT-HAL}& 59.1 & 58.4 & 57.2 & 59.4 \\
 &  {\small Video-MMMU} & 52.1 & 49.9 & 45.3 & 52.2\\
 \midrule
 \bottomrule
\end{tabular}
\caption{
Performance across different model sizes on VRIPT-HAL and VideoMMMU benchmarks.  {Base} denotes standard greedy decoding. SmartSight maintains its effectiveness and efficiency trade-off as model size increases. 
}
\label{tab:modelsize}
\end{table}

\begin{figure}[t!]
    \centering
    \includegraphics[width=0.95\linewidth]{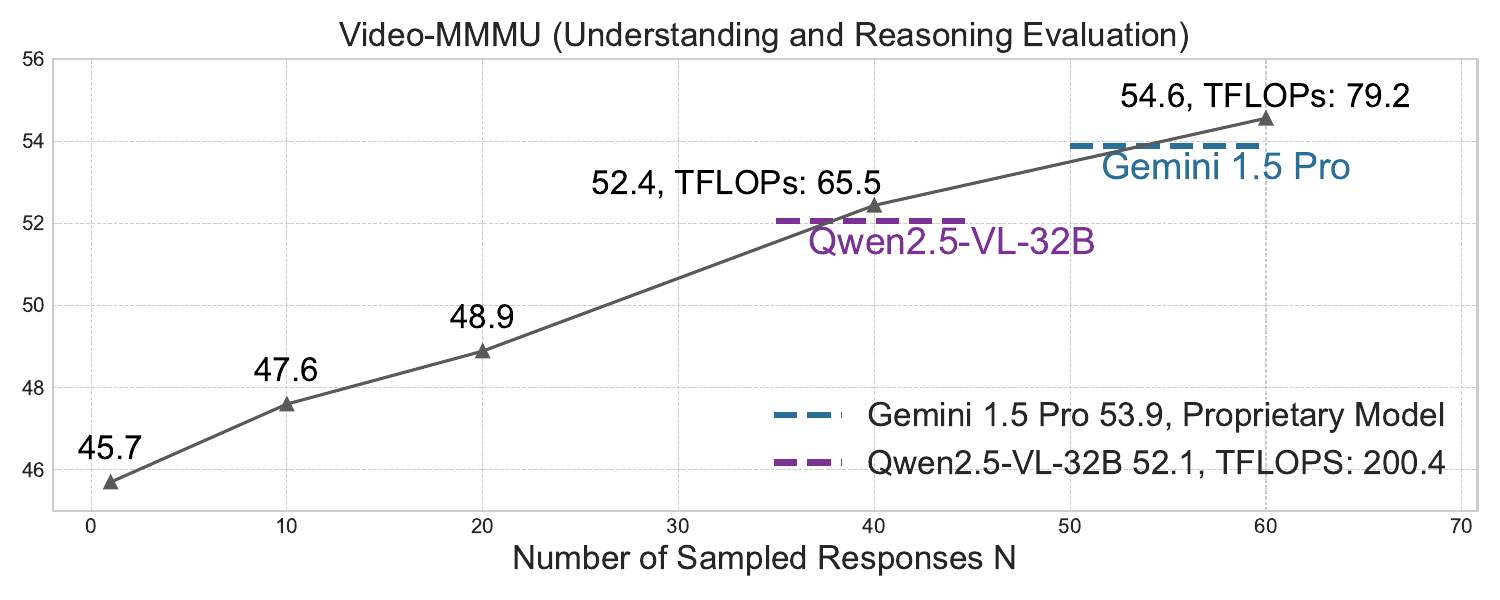}
    \includegraphics[width=0.95\linewidth]{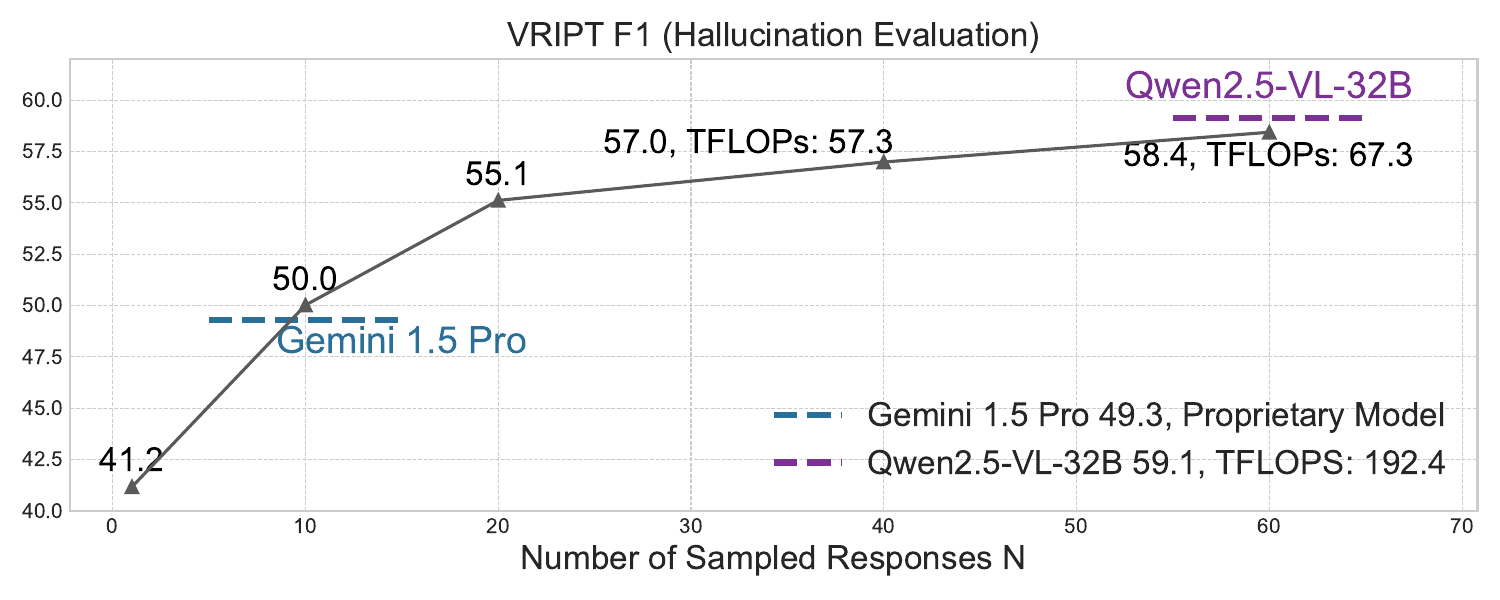}
    \caption{
Results of applying \name to Qwen2.5-VL-7B with different values of $N$. The figure shows that increasing $N$ enables test-time scaling, which is challenging  to achieve with prior methods. When $N = 60$, the 7B model achieves performance comparable to the Qwen2.5-VL-32B model and the proprietary Gemini 1.5 Pro.
    }
    \label{fig:scalen}
\end{figure}
\subsection{Main Results}
We evaluate our method on three representative Video-LLMs: LLaVA-NEXT-Video~\cite{li2025llavaonevision}, Qwen2.5-VL~\cite{qwen2.5vl}, and Video-R1~\cite{video-r1}.

\noindent \textbf{Effectiveness in Hallucination Mitigation.}
Our method achieves SOTA hallucination mitigation across both VQA-based and caption-based evaluations.
In contrast, existing approaches improve performance only on closed-ended VQA tasks, while exhibiting modest or even negative gains in open-ended captioning. 
Moreover, these methods typically fail to generalize across different Video-LLMs.

\begin{table}[t]
\centering
        \begin{tabular}{lr|rr}
\toprule
Model & Base & + Ours & Gain \\
\midrule
Gemini1.5 Pro & 49.3 & -  & - \\
Claude3.5-Sonnet & 44.6 & - &- \\
Video-LLaVA & 24.6 & 25.7 & 1.1 \\
LLaVA-OneVision-7B & 25.8 & 31.4 & 5.6 \\
InternVL3-8B & 30.4 & 34.1 & 3.7 \\
Qwen2.5-VL-7B & 47.9 & 50.0 & 2.1 \\
VideoChat-R1 & 48.4 & 52.1 & 3.7 \\
\bottomrule
\end{tabular}
\caption{\name is applied to a range of video-language models and consistently mitigates hallucinations on the VRIPT-HAL benchmark. Notably, several open-source models equipped with \name achieve performance on par with proprietary models.}
\label{tab:figure6}
\end{table}

\begin{figure}[t]
    \centering
    \includegraphics[width=1\linewidth]{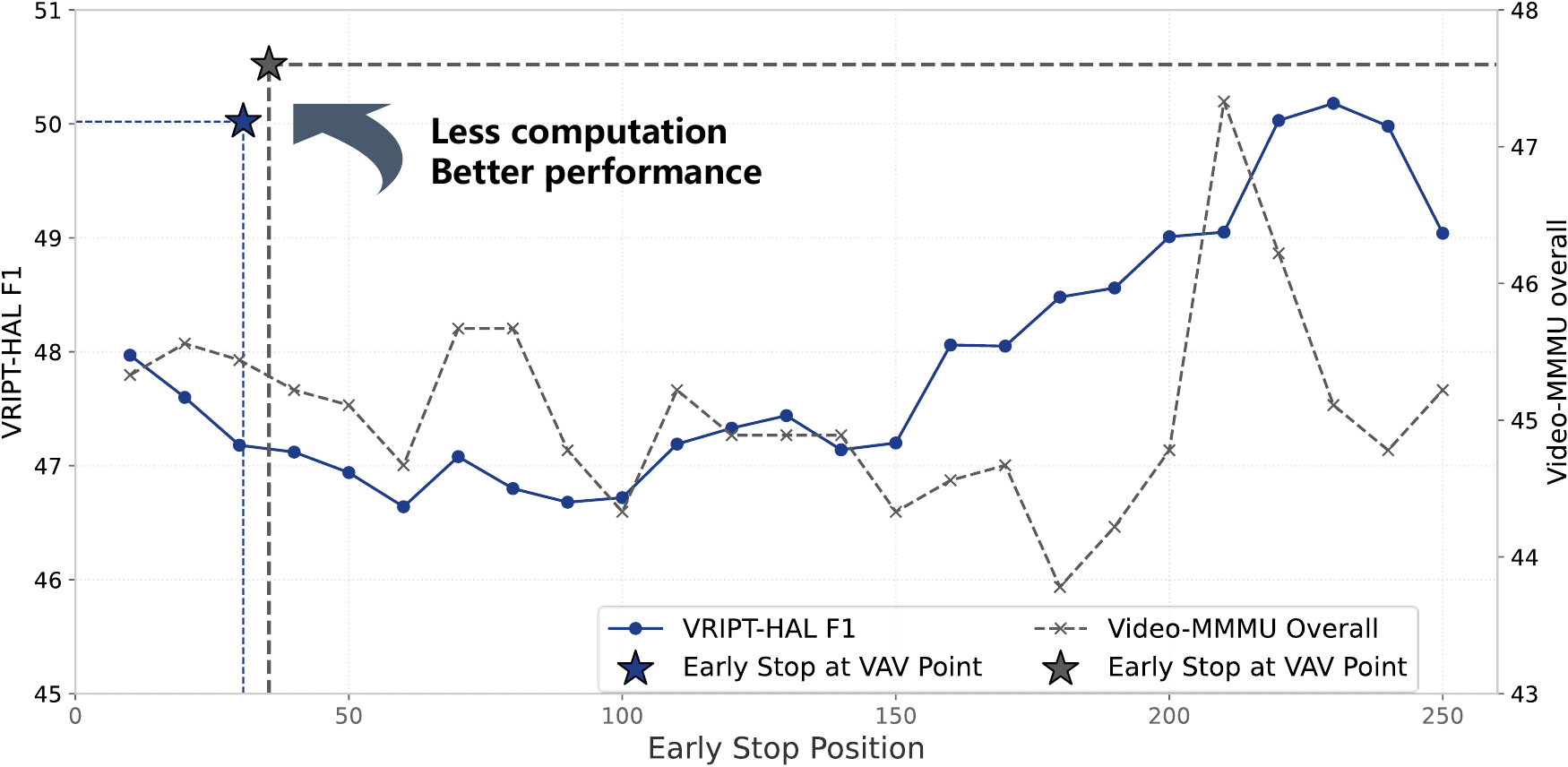}
    \caption{
Comparison of early stopping strategies. Blue and gray stars indicate the VAV early-stop points for VRIPT-HAL and Video-MMMU, respectively. The proposed VAV Point yields more accurate stopping decisions, resulting in superior hallucination suppression and reasoning performance compared to fixed-position stopping.
    }
    \label{fig:es}
\end{figure}
\noindent \textbf{Effectiveness in Complex Reasoning.}
On the challenging video understanding and reasoning benchmarks Video-MMMU and Video-MME, almost all baseline mitigation methods consistently degrade performance. Conversely, by leveraging reasoning paths that are more faithful to the video's content, our method improves the reasoning capabilities of nearly all evaluated Video-LLMs.

\noindent \textbf{Generalization on Different Video-LLMs.}
As shown in Table~\ref{tab:figure6}, \name demonstrates effective hallucination mitigation across various video-language models, including Video-LLaVA~\cite{lin2023video}, LLaVA-OneVision-7B~\cite{li2024llava}, and VideoChat-R1~\cite{li2025videochat}. Notably, several open-source models achieve performance on par with proprietary counterparts after applying \name. For example, VideoChat-R1 and Qwen2.5-VL-7B surpass Gemini 1.5 Pro. These results underscore the strong generalization capability of \name across diverse architectures.

\subsection{Computational and Scaling  Efficiency}
\noindent \textbf{Computational Efficiency.}
As shown in Table~\ref{tab:modelsize}, \name achieves efficiency by optimizing both the Prefill and Decode stages of Video-LLM inference. 
In the Prefill Stage, which accounts for a large portion of the total inference cost, \name processes only the original video input.
This design avoids the extra computation introduced by contrastive decoding methods, which typically forward both clean and perturbed inputs during this stage.
In the Decode Stage, where cost grows linearly with the number of generated tokens~\cite{kwon2023efficient}, SmartSight halts low-quality responses early and continues only the most promising one.
This strategy results in early termination at an average of 20.8\% and 12\% of the output length on Qwen2.5-VL-7B and Video-R1, respectively. Contrastive decoding methods, on the other hand, perform predictions at every decoding step using both clean and perturbed inputs, which results in substantially higher computational costs.

\noindent \textbf{Scaling with Model Size.} As shown in Table~\ref{tab:modelsize}, \name consistently  mitigates hallucination and improves understanding across different model sizes. Unlike competing methods, which are only effective on smaller models (e.g., 3B or 7B) and degrade at larger scales, SmartSight remains effective as model size increases, demonstrating favorable scalability.

\noindent \textbf{Scaling with the Number of Sampled Responses $N$.} Figure~\ref{fig:scalen} illustrates the effect of applying SmartSight to Qwen2.5-VL-7B under different numbers of sampled responses $N$. Increasing $N$ consistently improves understanding  on Video-MMMU and mitigates hallucination on VRIPT-HAL, demonstrating the ability of SmartSight to perform test-time scaling, which is difficult to achieve with existing methods.  Notably, when $N = 60$, Qwen2.5-VL-7B achieves performance comparable to the  Qwen2.5-VL-32B  while requiring significantly less computation, consuming 79.2 TFLOPs on VideoMMMU compared to 200.4 TFLOPs. It also performs competitively with the proprietary Gemini 1.5 Pro~\cite{team2024gemini}.

\subsection{Ablation Study}

\noindent \textbf{Effectiveness of the VAV Point.}
As shown in Figure~\ref{fig:es}, fixed stopping positions result in unstable performance on both VRIPT-HAL and Video-MMMU, highlighting the challenge of applying a one-size-fits-all truncation strategy. In contrast, early stopping at the VAV Point yields consistently better results, underscoring its effectiveness in precisely determining the optimal stopping position for each response.

\noindent \textbf{Component-wise Analysis.} 
As shown in Table~\ref{tab:abs_table}, we conduct a component-wise analysis on Qwen2.5-VL. 
Compared to the Baseline and Greedy settings, using either the Frame-level or Segment-level Collapse Score independently enables more accurate selection of high-quality responses.
Notably, early stopping via the VAV Point reduces decoding cost without  significantly  affecting performance.

\noindent \textbf{Ablations on Hyper-parameters.} 
As shown in Table~\ref{tab:alphaw}, increasing $\alpha$ from 1.0 to 1.2 and $w$ from 1 to 15 leads to consistent performance gains across both benchmarks, after which the metrics stabilize. These results suggest that SmartSight is robust to the choice of $\alpha$ and $w$.

\newcommand{\rotates}[2][0]{\rotatebox{#1}{#2}}
\newcommand{\rotatesf}[2][00]{\rotatebox{#1}{#2}}
\begin{table}[t!]
\setlength{\tabcolsep}{2pt} 
\small
\centering
\begin{tabular}{@{}lcccccc@{\hspace{1pt}}c@{}}
\toprule
& \textit{Greedy} & \textit{Baseline} & \textit{+FLC} & \textit{ +SLC }
& \textit{+TAC} & \textit{+TAC} & \textit{+ES} \\
\midrule
{VRIPT } & \multirow{2}{*}{47.9} & \multirow{2}{*}{41.2} & 48.2 & 48.3 & 50.1 & \multicolumn{2}{c}{50.0} \\
{-HAL}   & & & +7.0 & +7.1 & +8.9 & \multicolumn{2}{c}{+8.8} \\
\midrule
\rotatesf{Video} & \multirow{2}{*}{54.8} & \multirow{2}{*}{51.8} & 55.2 & 55.7 & 56.3 & \multicolumn{2}{c}{56.22} \\
{MME }        & & & +3.4 & +3.9 & +4.5 & \multicolumn{2}{c}{+4.4} \\
\bottomrule
\end{tabular}
\caption{Component-wise analysis on Qwen2.5-VL-7B. \textit{Greedy} uses greedy decoding without sampling, while \textit{Baseline} randomly selecting one response from $N=10$ sampled candidates. \textit{+FLC}, \textit{ +SLC }, and \textit{+TAC} select responses based on Frame-level, Segment-level, and TAC Scores, respectively.  \textit{+ES} applies early stopping at the VAV Point.}
\label{tab:abs_table}
\end{table}

\begin{table}[t]
\centering

\setlength{\tabcolsep}{2pt}
\small

\begin{tabular}{c|cccccccccc}
\toprule
$w$      & 1    & 3    & 5    & 7    & 9    & 11   & 13   & 15   & 17   & 19   \\
\midrule
VRIPT    & 48.6 & 49.7 & 50.0 & 50.1 & 50.5 & 50.1 & 50.2 & 50.3 & 50.1 & 50.1 \\
VMMMU    & 40.4 & 41.0 & 42.3 & 42.8 & 42.1 & 42.7 & 42.4 & 43.1 & 43.1 & 42.9 \\
\midrule
$\alpha$ & 1.0  & 1.1  & 1.2  & 1.3  & 1.4  & 1.5  & 1.6  & 1.7  & 1.8  & 1.9  \\
\midrule
VRIPT    & 47.3 & 49.2 & 50.0 & 49.7 & 49.2 & 48.8 & 48.4 & 47.8 & 48.6 & 48.7 \\
VMMMU    & 45.9 & 46.1 & 47.6 & 47.3 & 47.7 & 46.8 & 47.9 & 46.7 & 47.1 & 46.8 \\
\bottomrule
\end{tabular}
\caption{Performance comparison under different $w$ and $\alpha$ settings on VRIPT and VMMMU. VRIPT is short for VRIPT-HAL, and VMMMU is short for Video-MMMU.}
\label{tab:alphaw}
\end{table}

\section{Conclusion}
In this work, we present SmartSight, a training-free method that effectively mitigates perception hallucinations in Video-LLMs without compromising their understanding and reasoning abilities. By introducing the Temporal Attention Collapse score and the Visual Attention Vanishing Point, SmartSight not only suppresses hallucinated outputs but also achieves a favorable trade-off between computational efficiency and accuracy. Extensive experiments across ten diverse Video-LLMs validate its effectiveness, demonstrating notable gains in both hallucination reduction and video understanding performance. 
For the limitation, we  note that our method yields  modest improvements on short videos with simple scenes. This is primarily due to the fact that hallucinations are less likely to occur in such videos. Overall, these findings position SmartSight as a promising and generalizable solution for enhancing the reliability of Video-LLMs in more challenging and realistic settings.

\section*{Acknowledgements}
We are thankful to the shepherd and reviewers for their careful assessment and valuable suggestions, which have helped us improve this paper.
This work was supported in part by the National Natural Science Foundation of China (62472096, 62172104, 62172105, 62102093, 62102091, 62302101, 62202106).
Min Yang is a faculty of the Shanghai Institute of Intelligent Electronics \& Systems and Engineering Research Center of Cyber Security Auditing and Monitoring, Ministry of Education, China.

\bibliography{aaai2026}
\end{document}